%% file: main.tex
\documentclass[twocolumn]{reinpaper}
\input{paper-preamble}
\input{paper-info}
\begin{document}
\maketitle
\input{latex/abstract}
\input{latex/introduction}
\input{latex/background}
\input{latex/methodology}
\input{latex/evaluation}

\input{latex/conclusion}
\reinbibliography{adaagent}
\end{document}

%% file: paper-preamble.tex
\usepackage{makecell}

\newcommand{\takeaway}[2]{
\par\vfill\noindent
\begin{tikzpicture}
\node[
 draw=black, line width=0.9pt, fill=white,
 rounded corners=1.2pt,
 inner xsep=6pt, inner ysep=6pt,
 text width=\dimexpr0.97\columnwidth-12pt\relax,
 align=left, font=\small
] (tkbody) {\rule[-0.3ex]{0pt}{3.4ex}\ignorespaces #2};
\node[
 fill=black, text=white, font=\footnotesize\bfseries,
 inner xsep=6pt, inner ysep=2.5pt,
 anchor=south west
] (tkhead) at ([xshift=4pt,yshift=0.3pt]tkbody.north west) {\strut #1};
\end{tikzpicture}\par\newpage
}

\newcommand{\rev}[1]{#1}
\newcommand{\tao}[1]{#1}
\newcommand{\tool}{\texorpdfstring{A$^3$}{A3}}
\DeclareRobustCommand{\supfull}{\tikz[baseline=-0.55ex]\draw[fill=black](0,0)circle(0.55ex);}
\DeclareRobustCommand{\suppart}{\tikz[baseline=-0.55ex]{\draw(0,0)circle(0.55ex);\fill(0,0)--(0,0.55ex)arc(90:270:0.55ex)--cycle;}}
\DeclareRobustCommand{\supnone}{\tikz[baseline=-0.55ex]\draw(0,0)circle(0.55ex);}

\makeatletter
\patchcmd{\rein@titleblock}{\vspace{4pt}}{\vspace{2pt}}{}{\errmessage{A3 masthead spacing patch failed}}
\patchcmd{\rein@titleblock}{\vspace{6pt}}{\vspace{2pt}}{}{\errmessage{A3 rule spacing patch failed}}
\patchcmd{\rein@titleblock}{\vspace{10pt}}{\vspace{4pt}}{}{\errmessage{A3 title spacing patch failed}}
\patchcmd{\rein@titleblock}{\vspace{8pt}}{\vspace{4pt}}{}{\errmessage{A3 author spacing patch failed}}
\patchcmd{\rein@titleblock}{\vspace{4pt}{\small\rein@affiliation\par}}{\vspace{2pt}{\small\rein@affiliation\par}}{}{\errmessage{A3 affiliation spacing patch failed}}
\patchcmd{\rein@titleblock}{\vspace{10pt}}{\vspace{4pt}}{}{\errmessage{A3 abstract spacing patch failed}}
\makeatother

%% file: paper-info.tex
\title{\tool: Adaptive Audio Agent Framework for Collaborative Edge Computing}
\author{Zexin Li\textsuperscript{1}\reinorcid{0000-0001-8758-2151}\quad Xueyi Zhang\textsuperscript{2}\reinorcid{0009-0008-2394-3518}\quad
Tao Ren\textsuperscript{3}\reinorcid{0009-0007-9638-8386}\quad Yiming Chen\textsuperscript{4}\reinorcid{0009-0001-8557-1077}}
\reinaffiliation{\textsuperscript{1}Nanyang Technological University\quad
\textsuperscript{2}National University of Singapore\\
\textsuperscript{3}Rein Protocol Foundation\quad
\textsuperscript{4}ASUS Intelligent Cloud Services}
\reinemails{zexin.li@ntu.edu.sg\quad xueyi.zhang@nus.edu.sg\quad
tempest.ren@rein-protocol.org\quad MattYM\_Chen@asus.com}
\reinshorttitle{A3: Adaptive Audio Agents}
\reintrack{AUDIO AGENTS / EDGE COMPUTING}
\reinstatus{RESEARCH PREPRINT}
\reinkeywords{Audio agents; collaborative edge computing; hedged execution; local fallback}
\reingithub[github.com/ZexinLi0w0/A3]{https://github.com/ZexinLi0w0/A3}
\reinhidepartner
\makeatletter
\renewcommand{\rein@brandrow}{%
  \noindent\begin{minipage}[c]{.22\textwidth}\centering
    \includegraphics[width=.98\linewidth,height=31pt,keepaspectratio]{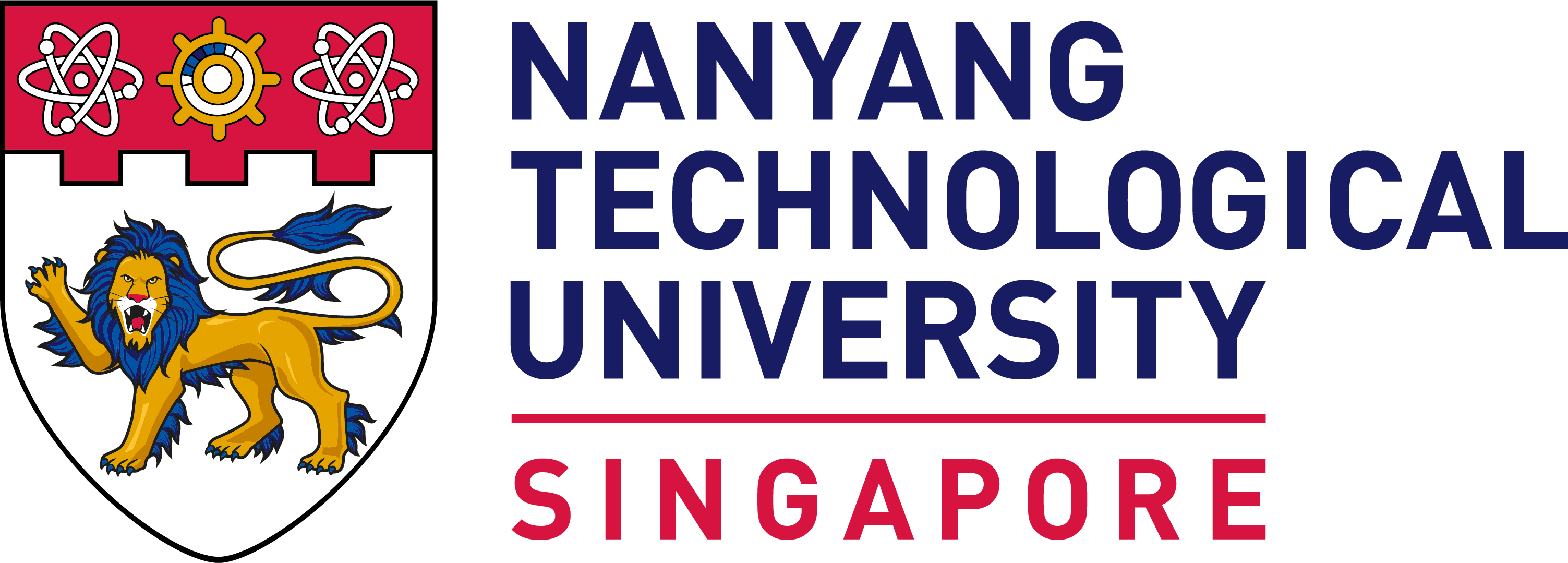}%
  \end{minipage}\hfill
  \begin{minipage}[c]{.22\textwidth}\centering
    \includegraphics[width=.98\linewidth,height=28pt,keepaspectratio]{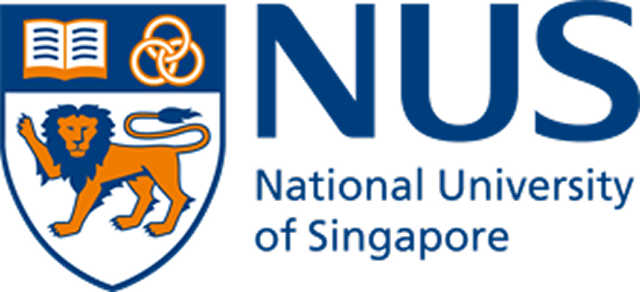}%
  \end{minipage}\hfill
  \begin{minipage}[c]{.22\textwidth}\centering
    \resizebox{.98\linewidth}{!}{\reinwordmark}%
  \end{minipage}\hfill
  \begin{minipage}[c]{.22\textwidth}\centering
    \includegraphics[width=.98\linewidth,height=28pt,keepaspectratio]{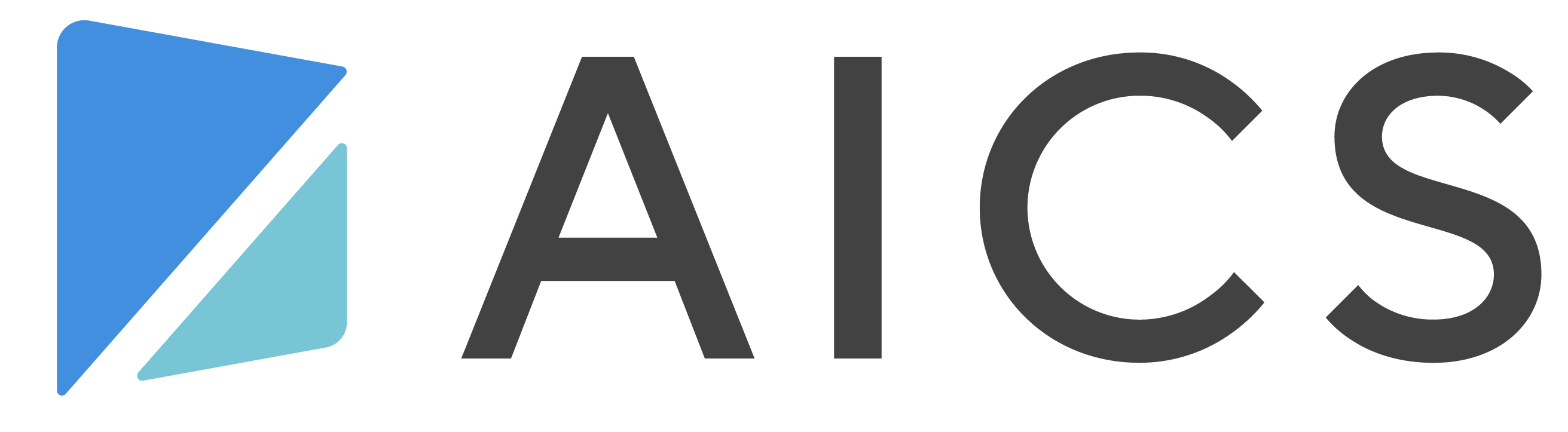}%
  \end{minipage}\par}
\makeatother

%% file: latex/abstract.tex
\begin{abstract}
Embedded audio agents need a nearby edge server for speed, but a server-dependent agent returns nothing when the link fails. We present \tool{}, an adaptive audio agent framework that places whole pipeline stages on the server, keeps a complete fallback agent on devices that can host one, and hedges both paths with a link-aware deadline that decides when to give up on the server. On Jetson devices and an RTX A4500 server, \tool{} matches the server-only agent on a healthy link, answers every request in a full outage on the AGX, and beats a percentile hedge in simulated outages.
\end{abstract}

%% file: latex/introduction.tex
\section{Introduction}

Audio agents let people interact with software by speaking~\cite{whisper2023,qwen2audio2024,llamaomni2025}. An audio agent chains automatic speech recognition (ASR), a large language model (LLM), and text-to-speech (TTS)~\cite{qwenaudioagent}. \rev{To stay available without a network,} audio agents increasingly run on embedded devices, such as robots~\cite{kim2024llmhri}, vehicles~\cite{cui2024drive}, and wearables~\cite{arakawa2024prism}, which also keeps the user's audio on the device. These devices have limited compute and memory~\cite{flexinfer2025,lessismore2025}, and a spoken question takes a long time to answer on them~\cite{rtlm2023,rtmobile2020}. Latency matters because a conversation feels natural only when replies are quick~\cite{stivers2009turn}, which a nearby edge server with a larger model can provide~\cite{kang2017neurosurgeon,mao2025easz}. \tao{Since a three-stage pipeline answers in seconds rather than within the sub-second gaps of human turn-taking, we treat latency as a cost to lower and an answer within a time limit as the requirement.}

Our fundamental research challenge is therefore the following.
\takeaway{Fundamental research challenge}{Under an uncertain and changing link, how much work should be placed on the remote resource, and when should it be abandoned, so that the remote resource speeds up the common case \rev{while the device keeps every request answerable}?}

While collaborative edge computing with a nearby server is attractive, the wireless link is unreliable. The link can add delay, limit bandwidth, and go down for seconds, and an agent that depends on the server then returns nothing. Running everything locally avoids the problem but pays the full device latency and needs larger hardware. Sharing the work between the device and the server therefore needs a rule for when to stop waiting for the link. Other fields have faced the same tension. Cloud-assisted driving keeps the on-board computation as a fallback~\cite{schafhalter2023leveraging}, and tail-tolerant services take the first reply of several replicas~\cite{dean2013tail}. Both assume a cheap fallback, while a three-model pipeline on a memory-limited device needs a fallback whose models are sized to that device.

The goal of this work is therefore an audio agent on an embedded device that is as fast as the server agent whenever the server is reachable, and responsive when it is not. Three design trade-offs follow from this challenge.

\noindent\textbf{What to run on the server.} Whole stages are a better unit than individual decoding steps. Collaborative inference splits work between the device and the server~\cite{kang2017neurosurgeon}, and for LLMs the natural split places a small draft model on the device and verification on the server~\cite{leviathan2023fast,chen2023accelerating}. We find that this split loses, since every step pays a network round trip. Whole stages avoid this cost and, when only text returns, depend less on the network, but they need a fallback that the device can hold.

\noindent\textbf{What the device can host.} The fallback must be a complete agent, but memory is scarce on embedded devices, so the fallback is tailored to each device, and a smaller model also answers less accurately. With a fallback of the right size, the remaining question is when to start it.

\noindent\textbf{When to fall back.} The answer depends on the state of the link. Starting the fallback costs the time already spent waiting plus the local latency, so under short outages waiting for the link finishes sooner, and a fallback that shares the device with other work slows that work down. A fallback that starts too early wastes compute and energy, and one that starts too late wastes the time already spent waiting.

We answer these three questions together with \tool, an \underline{a}daptive \underline{a}udio \underline{a}gent framework. \tool{} places whole pipeline stages on the server and returns audio or text according to the link profile. It keeps a tiered local agent tailored to the device and runs both paths as hedged execution, so the first complete reply wins and the other path is cancelled. A deterministic controller decides when the local agent starts from the observed link state and the server latency of requests that ran alone on the device.

We evaluate \tool{} on a Jetson Orin AGX, an Orin Nano, and a server with RTX A4500 GPU, under emulated jitter, bandwidth limits, loss, and outages. On a healthy link, \tool{} matches the server-only agent and keeps the fallback idle, and in a full outage it answers every request while the server-only agent stays silent. \rev{\tool{} combines a complete on-device fallback, whole-stage server execution, and a link-aware fallback deadline in one audio framework, and we compare it in simulation with a percentile hedge under the same fallback cap.} In simulation, \tool{} completes 0.91 of the requests within 30\,s against 0.87 for the percentile hedge, and in a replay of a field trip on the Singapore MRT it answers 73 of 73 requests where the server-only agent answers 70. In this work, we make the following major contributions:
\begin{itemize}[leftmargin=*,itemsep=0pt,topsep=2pt]
\item We propose \tool, a framework with a tiered local agent, a server path that adapts what it returns to the link, and a hedged executor \rev{that answers whenever a path completes within a time limit}.
\item We design a deterministic, link-aware deadline for starting the fallback, whose latency estimate excludes requests that shared the device with a fallback.
\item We evaluate \tool{} with six models (two ASR models, three Qwen language models, and one TTS model) on two Jetson devices and an A4500 server under \tao{twelve network conditions (eight emulated link conditions and four new outage processes).} \tool{} answers every request in a full outage \rev{on the AGX} where the server-only agent answers none, it halves the latency of the device alone on a healthy link, it lowers the latency under a bandwidth cap from 17.6\,s to 9.7\,s, and on four new outage processes it completes 12.5 percentage points more requests on time than waiting on average, and up to 35 percentage points in the best case.
\end{itemize}

%% file: latex/background.tex
\section{Background and Motivation}
\label{sec:bg}


\subsection{Audio agents and related work}
\begin{table}[t]
\centering
\caption{Comparison with the closest methods. \supfull, \suppart, and \supnone{} denote support, partial support, and no support. Link-aware: the decision uses the measured link state. Deadline: the method targets a completion deadline.}
\label{tab:related}
\footnotesize
\setlength{\tabcolsep}{2.5pt}
\begin{tabular}{@{}lcccccc@{}}
\toprule
\textbf{Method} & \textbf{LLM} & \textbf{\makecell{Server\\offload}} & \textbf{\makecell{Local\\fallback}} & \textbf{\makecell{Link-\\aware}} & \textbf{Deadline} & \textbf{Audio} \\
\midrule
Neurosurgeon~\cite{kang2017neurosurgeon} & \supnone & \supfull & \suppart & \supfull & \supnone & \suppart \\
Hedged requests~\cite{dean2013tail} & \supnone & \supnone & \supnone & \supnone & \suppart & \supnone \\
Cloud + on-board~\cite{schafhalter2023leveraging} & \supnone & \supfull & \supfull & \suppart & \supfull & \supnone \\
EdgeShard~\cite{zhang2025edgeshard} & \supfull & \supfull & \supnone & \suppart & \supnone & \supnone \\
Hybrid SLM/LLM~\cite{hao2024hybrid} & \supfull & \supfull & \suppart & \supnone & \supnone & \supnone \\
HCInfer~\cite{liu2025hcinfer} & \supfull & \supnone & \supnone & \supnone & \supfull & \supnone \\
\midrule
\textbf{\tool{}} & \supfull & \supfull & \supfull & \supfull & \suppart & \supfull \\
\bottomrule
\end{tabular}
\end{table}
An audio agent turns speech into a reply through ASR, an LLM, and TTS, and we build on the Qwen Audio Agent runtime~\cite{qwenaudioagent} with Qwen3-ASR, Qwen3~\cite{yang2025qwen3}, Kokoro-82M~\cite{kokoro82m}, and vLLM~\cite{kwon2023vllm}. Speech and audio language models such as Whisper~\cite{whisper2023}, SALMONN~\cite{salmonn2024}, Qwen-Audio~\cite{qwenaudio2023,qwen2audio2024}, LLaMA-Omni~\cite{llamaomni2025}, and multi-audio models~\cite{chen2024beyond} assume server hardware, and Dong et al.~\cite{rtmobile2020} and Paramanayakam et al.~\cite{lessismore2025} accelerate speech recognition and function calling on the device only, and state space models speed up long-context speech recognition~\cite{gao2024speechmamba}.

Collaborative edge computing partitions a model between a device and a server~\cite{kang2017neurosurgeon,mao2025easz,hidp2025,pagliari2020}, shards LLMs across edge nodes~\cite{zhang2025edgeshard}, routes between a small and a large model~\cite{hao2024hybrid}, and schedules inference under deadlines~\cite{liu2025hcinfer,pnc2023,edgert2022,rtlm2023,gao2025rtss}, but it assumes that the link stays up. Hedged requests~\cite{dean2013tail} start a delayed second request at a latency percentile and cancel the slower one, and cloud execution with an on-board fallback~\cite{schafhalter2023leveraging} handles failures in driving. \tool{} builds on both and adds the return form, an outage-conditioned deadline, a fallback budget, and a complete audio pipeline on a memory-limited device.

\begin{figure}[t]
\centering
\includegraphics[width=\columnwidth]{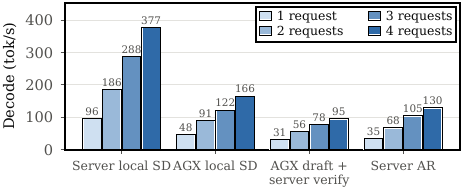}
\caption{Qwen3-4B decode throughput by placement and concurrency on the A4500. SD: speculative decoding; AR: autoregressive decoding; tok/s: tokens/s.}
\label{fig:e8}
\end{figure}

\subsection{Motivational case study}
\label{sec:motivation}
The case study uses the testbed of Sec.~\ref{sec:eval} (an AGX, a Nano, and an RTX A4500 server over Wi-Fi 6), Qwen3-4B on the server, a smaller Qwen3 model with a draft model on the devices, and four synthetic spoken questions. TTS uses Kokoro~\cite{kokoro82m}, an 82M-parameter model.

\noindent\textbf{Observation 1: TTS dominates on the CPU and nearly disappears on the GPU.} The cost of TTS is a known concern~\cite{gao2025ttslow}, and Kokoro takes more than 40\,s per reply on the AGX CPU but about 2\,s on its GPU, where it needs 0.11\,s on the server, 0.25\,s on the AGX, and 0.79\,s on the Nano per 10\,s of audio. With TTS on the GPU, the LLM becomes the largest remaining cost, with 4.2\,s of the 7.9\,s stage time on the AGX, 11.2\,s of 17.1\,s on the Nano, and 2.3\,s of 4.2\,s on the server, so server execution gains most on the LLM stage.

\noindent\textbf{Observation 2: splitting speculative decoding slows decoding.} For one request, speculative decoding on the server (Server SD) reaches 95.8\,tok/s, on the AGX (AGX SD) 47.9\,tok/s, autoregressive decoding on the server (AR) 35.4\,tok/s, and the split with the draft model on the AGX and the verification on the server (Split) only 31.1\,tok/s; with four concurrent requests the four placements reach 377, 166, 130, and 95\,tok/s in the same order (Fig.~\ref{fig:e8}), so Split is slower than the AGX alone and reaches about one third of Server SD with one request and one quarter with four. Each decoding step needs the draft tokens from the AGX and the verification result from the server, so every step adds a network round trip and the two GPUs wait for each other, and a pipelined split still falls behind as concurrency grows. These measurements favor whole-stage placement, which requires the device to host a small pipeline.

\noindent\textbf{Observation 3: the smallest device lacks the memory for the whole pipeline.} \tao{Measured one at a time, the three services add 7.1\,GB, or 93\% of the 7.6\,GB of the Orin Nano, against 37\% of the memory of the AGX and 78\% of the GPU memory of the server. With the 1.6\,GB that the system already uses, they exceed the Nano.} Both server paths run ASR remotely, so the audio leaves the device, and only a fallback that runs locally keeps it there. Each service runs in its own process with its own CUDA context and cache on one shared pool, so a local pipeline on the Nano needs a smaller model or all three services must share one process.

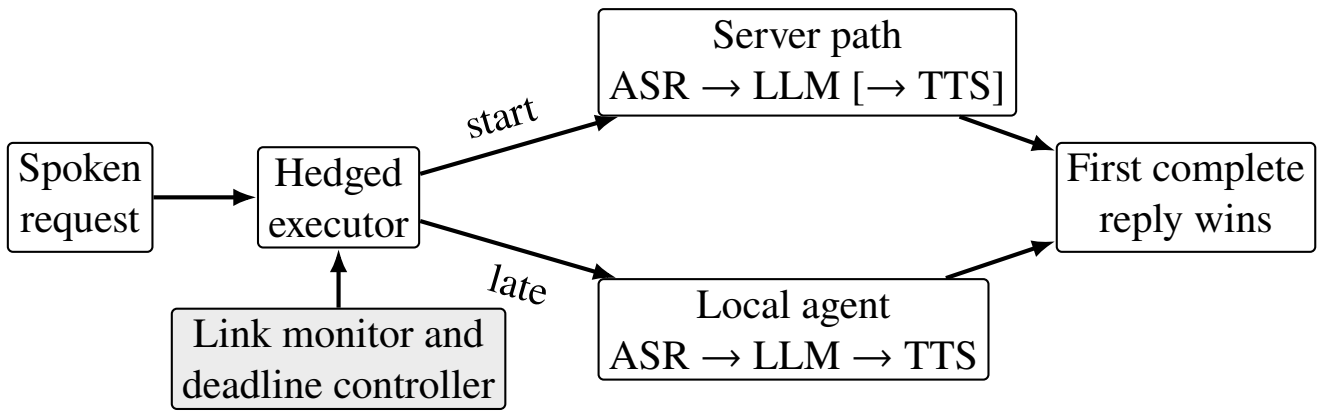
\begin{figure}[t]
\centering
\resizebox{\columnwidth}{!}{%
\begin{tikzpicture}[font=\footnotesize,node distance=4mm and 7mm,
  box/.style={draw,rounded corners=1pt,minimum height=6mm,align=center,inner sep=2pt},
  arr/.style={-{Latex[length=1.6mm]},thick}]
\node[box] (mic) {Spoken\\request};
\node[box,right=of mic] (exec) {Hedged\\executor};
\node[box,above right=2mm and 12mm of exec,minimum width=26mm] (rem) {Server path\\ASR $\to$ LLM [$\to$ TTS]};
\node[box,below right=2mm and 12mm of exec,minimum width=26mm] (loc) {Local agent\\ASR $\to$ LLM $\to$ TTS};
\node[box,right=of exec,xshift=36mm] (out) {First complete\\reply wins};
\node[box,below=4mm of exec,fill=gray!15] (ctl) {Link monitor and\\deadline controller};
\draw[arr] (mic) -- (exec);
\draw[arr] (exec) -- node[above,sloped,pos=.45]{start} (rem);
\draw[arr] (exec) -- node[below,sloped,pos=.55]{late} (loc);
\draw[arr] (rem) -- (out);\draw[arr] (loc) -- (out);
\draw[arr] (ctl) -- (exec);
\end{tikzpicture}}
\caption{\tool{} overview. The server path starts first, the local agent starts after a deadline chosen by the controller, and the first complete reply is played while the other path is cancelled.}
\label{fig:overview}
\end{figure}

\begin{algorithm}[t]
\footnotesize
\caption{Link-aware hedged execution of one request}
\label{alg:aaa}
\begin{algorithmic}[1]
\Require outage history $\mathcal{O}$, recent server latencies $\mathcal{W}$ from requests running alone on the device, estimated local latency $L$, link state $u$ (1 up, 0 down), response limit $H$, multiplier $c$, margin $\delta$, floor $D_{\min}$, fallback concurrency cap $B$
\Ensure first complete reply, or an unanswered request after $H$
\State reply form $f \gets$ text \textbf{if} returning audio takes longer than synthesizing it on the device \textbf{else} audio
\State $S \gets \mathrm{median}(\mathcal{W})$; $\hat{S} \gets$ 95th percentile of $\mathcal{W}$; $R \gets$ mean remaining time of the past outages in $\mathcal{O}$ that outlast the current one
\State $D \gets$ the deadline of Eq.~(\ref{eq:deadline}) from $u$, $S$, $\hat{S}$, $R$, and $L$
\State start server path with $f$ and elapsed-time clock $t$ at 0
\While{no reply \textbf{and} $t < H$}
    \State \textbf{if} a local agent exists and has not started \textbf{and} ($t \ge D$ \textbf{or} the server path failed) \textbf{then} start it when fewer than $B$ fallbacks are running
\EndWhile
\State \textbf{if} no reply \textbf{then} cancel both paths and report the request as unanswered
\State play the first complete reply and cancel the other path
\State \textbf{if} the server path won \textbf{and} the local agent never started \textbf{then} add its latency to $\mathcal{W}$
\end{algorithmic}
\end{algorithm}

\begin{figure}[t]
\centering
\includegraphics[width=0.455\columnwidth]{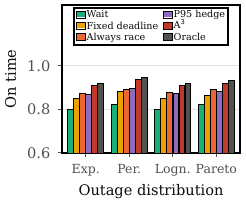}\hfill
\includegraphics[width=0.455\columnwidth]{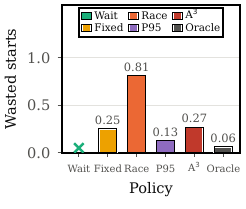}
\caption{Simulated policy comparison. P95: 95th-percentile hedge; Exp., Per., Logn.: exponential, periodic, lognormal outages.}
\label{fig:e3sim}
\vspace{6pt}
\includegraphics[width=0.455\columnwidth]{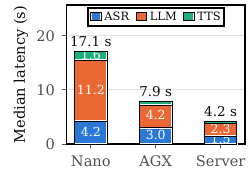}\hfill
\includegraphics[width=0.455\columnwidth]{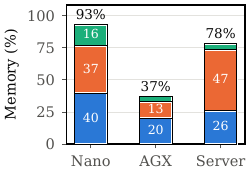}
\caption{Median stage latency (left) and memory use as a share of the device capacity (right) of each service run alone.}
\label{fig:e0}
\end{figure}

\begin{figure}[t]
\centering
\includegraphics[width=0.455\columnwidth]{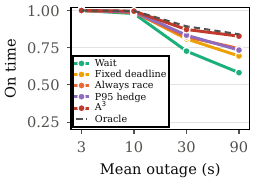}\hfill
\includegraphics[width=0.455\columnwidth]{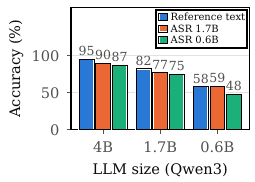}
\caption{Simulated share of requests completed within 30\,s against the mean outage length at a mean uptime of 60\,s (left), and end-to-end accuracy on 100 spoken GSM8K questions by LLM size and input (right).}
\label{fig:e3}\label{fig:e6}
\vspace{6pt}
\includegraphics[width=0.455\columnwidth]{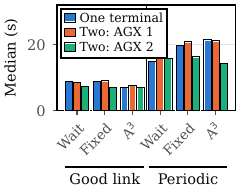}\hfill
\includegraphics[width=0.455\columnwidth]{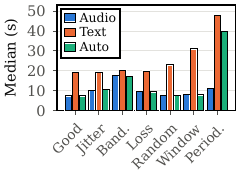}
\caption{Shared server (left); Nano paths (right). Auto: adaptive audio/text reply; Band.: bandwidth cap; Period.: periodic outages.}
\label{fig:e5}\label{fig:nano}
\end{figure}

\begin{figure}[t]
\centering
\includegraphics[width=\columnwidth]{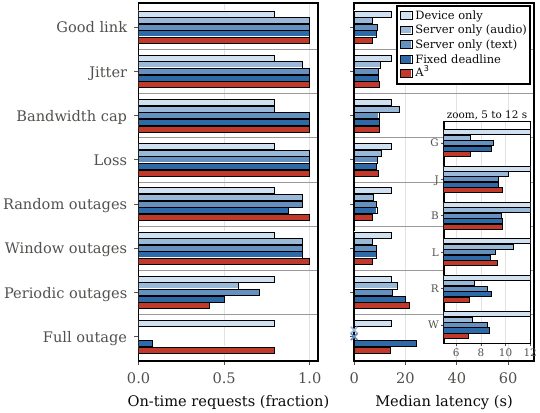}
\caption{AGX: fraction answered within 20\,s (left), median latency (right). Inset: G good link, J jitter, B bandwidth cap, L loss, R random, W window outages; crosses: unanswered requests.}
\label{fig:robust}
\end{figure}

\begin{figure}[t]
\centering
\includegraphics[width=0.455\columnwidth]{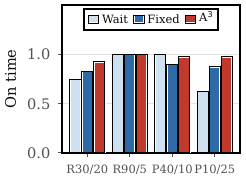}\hfill
\includegraphics[width=0.455\columnwidth]{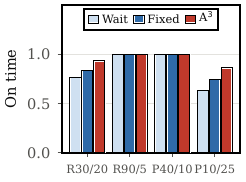}
\caption{New outages: testbed (left), simulator (right). R/P: random/periodic; numbers: mean up/down times (s).}
\label{fig:e3v}
\vspace{6pt}
\centering
\includegraphics[width=0.455\columnwidth]{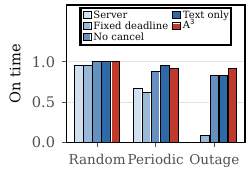}\hfill
\includegraphics[width=0.455\columnwidth]{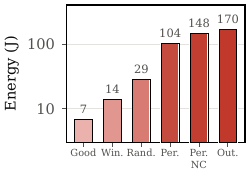}
\caption{Controller ablation (left); fallback energy, log scale (right). Win., Rand., Per., Out.: window, random, periodic, full outage; NC: no cancellation.}
\label{fig:e7}\label{fig:e4}
\end{figure}

\begin{figure}[t]
\centering
\includegraphics[width=0.53\columnwidth]{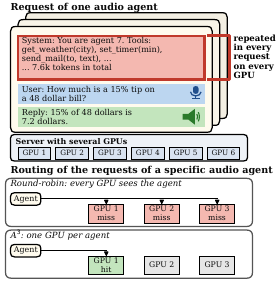}\hspace{0.3mm}\includegraphics[width=0.46\columnwidth]{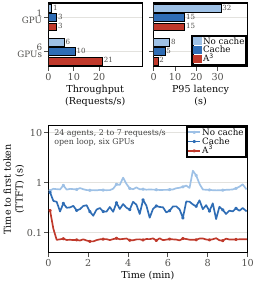}
\caption{Scale-up agent: one request and its routing over GPUs (left), throughput, 95th percentile latency, and time to first token (right).}
\label{fig:tool}
\end{figure}

\begin{figure}[t]
\centering
\includegraphics[width=\columnwidth]{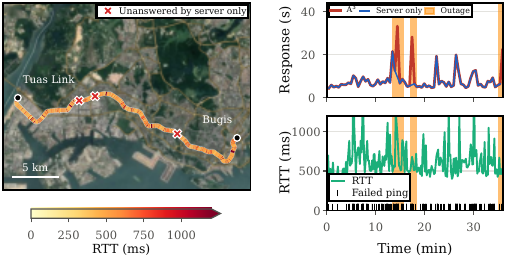}
\caption{Field study from Tuas Link to Bugis: round-trip time (RTT) along the route (left), response time and RTT over time (right).}
\label{fig:field}
\end{figure}

%% file: latex/methodology.tex
\section{\tool{} Design}
\label{sec:method}
\subsection{Overview}
We propose \tool{}, which answers each request with hedged executions of a server path and a local agent, and it runs on the device next to the audio agent gateway. The executor starts the server path, adds the local agent after a deadline, and plays the first complete reply, and cancels the other path (Fig.~\ref{fig:overview}). 

\noindent\textbf{Objective.} For $N$ requests $i=1,\ldots,N$ and response limit $H$, completion is $C_i=\min(T_{s,i},\,F_i+Q_i+L_{c,i})$ (unanswered if $C_i>H$), where $T_{s,i}$ is the server-path completion time for reply form $f_i$ (audio/text), $F_i$ the fallback trigger time (deadline $D_i$ or earlier server failure), $Q_i$ the wait for a fallback slot, and $L_{c,i}$ the local duration under contention. Completion and trigger times are relative to arrival. Let $\mathbf{1}[\cdot]$ be 1 if true, 0 otherwise. For target time $T$, the miss rate (late or unanswered) is $M=\frac{1}{N}\sum_i\mathbf{1}[C_i>T]$ and waste is $\bar W=\frac{1}{N}\sum_i W_i$, where $W_i=\mathbf{1}[F_i<T_{s,i}\le F_i+Q_i+L_{c,i}]$ marks a fallback that starts and loses to the server. With miss budget $\epsilon$, \tool{} solves
\begin{equation}
\min_{f,\,D}\ \bar W(f,D)\quad\text{s.t.}\quad M(f,D)\le\epsilon,\ \ n(t)\le B\ \ \forall t,
\end{equation}
where $f,D$ are reply-form/deadline choices with request-$i$ values $f_i,D_i$, $n(t)$ counts fallbacks running at time $t$, and $B$ caps concurrency. Shrinking $D$ lowers misses but raises waste, so the solution is the largest feasible $D$. The online rules approximate this choice from link state because server latency and outages are unknown. \tao{They do not use $\epsilon$; Sec.~\ref{sec:eval} reports achieved miss rate and waste. Monotonicity in $D$ assumes $B$ rarely delays fallback.}

\subsection{Components}
\noindent\textbf{Tiered local agent.} The local agent is a complete ASR, LLM, and TTS pipeline that is sized to the device. On the smallest device we use, the memory next to the operating system supports only the server path, so a full outage leaves its requests unanswered.

\noindent\textbf{Server path that adapts to the link.} The server path runs whole stages in one of two forms. In the full form, ASR, the LLM, and TTS run on the server and the reply audio crosses the link, and in the text form only ASR and the LLM run on the server and TTS runs on the device, which saves downlink bandwidth. The controller returns text when the estimated audio download time, from the measured downlink bandwidth, exceeds the device TTS time.

\noindent\textbf{Adaptive deadline.}
\label{sec:adaptive}
The deterministic controller sets fallback deadline $D$ from link state $u$ (1 up, 0 down), probed every second (down after 2 failures), using latency multiplier $c=1.5$, margin $\delta=1$\,s, and floor $D_{\min}=4$\,s:
\begin{equation}
D=\begin{cases}
\max\big(D_{\min},\ c\,\hat{S}\big) & u=1,\\
0 & u=0 \text{ and } L<R+S-\delta,\\
R+c\,S & u=0 \text{ and } L\ge R+S-\delta,
\end{cases}
\label{eq:deadline}
\end{equation}
where $\mathcal{W}$ holds the last 40 server latencies of requests running alone on the device, $\hat{S},S$ are its 95th percentile and median ($S=7$\,s initially), $L$ estimates local latency, and $R$ is the mean residual of past outages longer than the current one (30\,s without history). On healthy links, $D>\hat{S}$, so \tao{if $\mathcal{W}$ represents current latency,} at most \tao{about} 5\% start a fallback; $D_{\min}$ remains below the fastest measured server replies (about 6\,s). During outages, immediate fallback completes at $L$ versus about $R+S$ for waiting; $\delta$ covers losing-fallback cost, yielding immediate start if $L<R+S-\delta$, otherwise at $R+cS$. Only requests running alone update $\mathcal{W}$ because concurrent fallbacks slow the server path. Replies enter $\mathcal{W}$ only before fallback starts, so censoring at $D$ can underestimate $\hat{S}$; updates stop if every request starts a fallback.

\rev{\noindent\textbf{Variant with prior and budget.} A second variant adds prior outages of 10, 20, and 40\,s, leaves the down state after 3 successful probes, and changes reply form only for at least 15\% improvement. Let $T_s,Q,L_c$ estimate $T_{s,i},Q_i,L_{c,i}$ for the current request. It picks $D=\arg\min_{d\in\mathcal{D}}\mathbb{E}_{r}[\min(T_s,\,d+Q+L_c)]$, where $d$ is a candidate deadline in $\mathcal{D}=\{0,2,4,6,8,10,14,20,\infty\}$\,s ($\infty$ disables fallback), and $\mathbb{E}_{r}$ averages over residual outage samples $r$. Switching requires a gain of at least 1\,s. It models each concurrent fallback as 50\% local slowdown; the cap $B=3$ gives $Q>0$ when all slots are busy.} Below, the basic controller follows Eq.~(\ref{eq:deadline}).

\noindent\textbf{Cancellation and fallback budget.} When one path completes, the executor cancels the other, which closes its request and stops the local GPU work \rev{and signals the server to stop, which an outage delays}. The budget $B$ limits the local agents that run at the same time.

\subsection{Algorithm}
Algorithm~\ref{alg:aaa} gives the basic controller with per-request time complexity $O(|\mathcal{W}|+|\mathcal{O}|+H/\Delta)$, where $\mathcal{O}$ is bounded outage history, $|\cdot|$ counts entries, and $\Delta$ is the polling interval.

%% file: latex/evaluation.tex
\section{Evaluation}
\label{sec:eval}

\subsection{Setup}
\noindent\textbf{Platforms.} The terminals are a Jetson Orin AGX with 32\,GB of memory and an Orin Nano with 8\,GB. The server is one RTX A4500 of an eight-GPU machine, which the terminals reach over Wi-Fi 6, and every service runs on the GPU of the machine it belongs to.

\noindent\textbf{Models.} The AGX fallback runs Qwen3-ASR-1.7B and Qwen3-1.7B~\cite{yang2025qwen3} with a DSpark draft model~\cite{leviathan2023fast,chen2023accelerating}, the server runs Qwen3-ASR-1.7B and Qwen3-4B with its draft model, and both sides use Kokoro-82M~\cite{kokoro82m} for TTS, served with vLLM~\cite{kwon2023vllm}.

\noindent\textbf{Workload.} Poisson arrivals through the real gateway use four synthetic spoken questions: 0.05 requests/s on the AGX, 0.1 on the Nano, with 24 measured requests per cell (20 per seed on new outage processes). Latency is end to end. Targets are $T=30$\,s for simulation and new outage processes, $T=20$\,s elsewhere; requests without a reply by $H=45$\,s are unanswered. \tao{With 24 requests, a cell has a 95\% Wilson interval of about $\pm0.18$ at 0.67, so we read testbed differences below this width as trends and rely on the simulator for the ranking.} \tool{} is the basic controller in Figs.~\ref{fig:e5} to~\ref{fig:e3v} and the energy chart of Fig.~\ref{fig:e7}, and the variant with prior and budget in Fig.~\ref{fig:e3sim} and the left chart of Fig.~\ref{fig:e7}.

\noindent\textbf{Network.} A user-space proxy before the server adds delay and jitter, caps bandwidth, stalls chunks to mimic loss, or interrupts the link. The interruptions are periodic, random, one window, or a full outage.

\noindent\textbf{Baselines.} We compare \tool{} with the policies of the closest methods (Table~\ref{tab:related}): waiting for the server, as static offloading does~\cite{kang2017neurosurgeon,zhang2025edgeshard}; a fixed deadline of 10\,s for the on-board fallback~\cite{schafhalter2023leveraging}; always racing both paths; and the percentile hedge~\cite{dean2013tail}, a second request after a static 95th percentile of the calibration latencies of the server, without link state, under the same cap of three concurrent fallbacks. The simulator adds an oracle that knows the outage schedule, and the testbed runs the device alone, the server with audio, the server with text, the fixed deadline, and \tool{}. In the simulator, all policies see the same arrivals and the same outage schedules.

\subsection{Overall Effectiveness}
\noindent\textbf{Deadline policy in simulation.} The discrete-event simulator draws the server and local latencies from the measured samples, and the link follows an on and off outage process. A started fallback runs to completion and slows by an assumed 50\% for each other fallback that runs at the same time, and the oracle is a schedule-aware reference without contention. Fig.~\ref{fig:e3sim} shows the share of requests completed within 30\,s, averaged over twelve outage conditions, five seeds of 600 requests each, and four outage length distributions (left), and the share of requests whose fallback starts but loses to the server in the exponential case (right). In the exponential case, \tool{} completes 0.91 against 0.80 for waiting, 0.85 for a fixed deadline, 0.87 for the percentile hedge, and 0.88 for always racing, close to the 0.92 of the oracle, and the other distributions give the same order with 0.91 to 0.94 for \tool{}. \tool{} starts a fallback for 49\% of the requests and loses 27\% to the server, against 25\% for a fixed deadline, 13\% for the percentile hedge, and 81\% for always racing. The margin over the percentile hedge is 5.1 points without slowdown and 4.0 points at the assumed 50\% (paired 95\% bootstrap interval 2.9 to 5.2), and it reaches 0 at a slowdown of 100\% and at 0.5 requests per second, where every policy saturates.

\noindent\textbf{Stage costs.} Fig.~\ref{fig:e0} gives the latency and the memory of each service run alone, with the largest model that fits each device. The AGX needs 7.9\,s per request, 1.9$\times$ the 4.2\,s of the server, and the Nano needs 17.1\,s, 2.2$\times$ the AGX, so it relies on the server for all three stages of the pipeline.

\noindent\textbf{Devices that rely on the server path.} A device without a resident fallback depends on the server, as the Nano shows on the right of Fig.~\ref{fig:nano}, where a cross marks each of its three server paths that answers no request in a full outage. Under periodic outages Auto takes 40.0\,s against 11.2\,s for audio, because the slow Nano TTS dominates the text returned during the down phase, and only 0.13 of its requests finish within 20\,s against 0.80. \tao{The form rule estimates the download time from the downlink measured during the down phase, which collapses, so Auto picks text exactly when the slow Nano TTS makes it worse; conditioning the form on $u$ avoids this case, and on the AGX the rule runs inside \tool{}.}

\noindent\textbf{Availability under a full outage.} Both server-only paths answer none of the 24 requests, while \tool{} answers all 24 on the AGX with a median of 14.0\,s, close to the 13.9\,s of the oracle (Fig.~\ref{fig:robust}), and a fixed deadline answers all of them at 24.4\,s with 0.08 within 20\,s against 0.79. Once the probes mark the link down, \tool{} starts the local agent at once for every request, so its median is close to the 14.6\,s of the device alone. Its 90th percentile is 26.7\,s and its slowest reply 32.5\,s, against 36.7\,s and 42.7\,s for the fixed deadline, which waits 10\,s before every fallback, and 0.96 of its replies arrive within 30\,s against 0.79.

\noindent\textbf{Placement and robustness on the testbed.} On a good link the server path takes 7.2\,s with audio and 9.0\,s with text on the Orin AGX, against 14.6\,s on the device, and on the Nano audio takes 7.4\,s against 19.1\,s for text because its own TTS is slow, which Auto matches. \tool{} keeps the fallback idle on a healthy link and matches the server path, with a median of 7.1\,s against 7.2\,s on a good link and 9.4\,s against 10.6\,s under loss, since its estimate uses only requests that ran alone\tao{, whereas an estimate that includes requests that overlapped a fallback doubles it to 15.4\,s.} Under a bandwidth cap, audio takes 17.6\,s (0.79 within 20\,s) and text 9.6\,s, and \tool{} selects text, which gives 9.7\,s, all on time. Under periodic outages our first controller took 21.5\,s with 0.42 of the requests within 20\,s because its fallbacks slowed each other, and the variant with prior and budget reached 15.4\,s and 0.67 on the first AGX board, against 13.3\,s for the oracle, and two terminals on one server match one terminal (Fig.~\ref{fig:e5}, left).

\noindent\textbf{Validation on new outage processes.} Fig.~\ref{fig:e3v} shows the share of requests completed within 30\,s on the testbed, with 40 requests per cell pooled over two seeds that share one outage schedule, and in the simulator (R random, P periodic, with the mean up and down time in seconds). \tool{} leads on the two processes with long outages, with 0.93 against 0.75 for waiting under random outages of 30\,s up and 20\,s down, and 0.97 against 0.62 under periodic outages of 10\,s up and 25\,s down, and the simulator follows the same order within 0.13 of the testbed. Averaged over the four processes, \tool{} completes 12.5 points more requests than waiting on the testbed.

\subsection{Ablation Study}
\noindent\textbf{Controller components.} Fig.~\ref{fig:e7} ablates the controller on a shared AGX with 24 requests per cell, and its left chart shows the share of on-time requests for five variants, where No cancel and Text only use the basic controller and A$^3$ is the variant with prior and budget. Under periodic outages A$^3$ is on time for 0.92 of the requests \tao{on the second AGX board (Wilson interval 0.74 to 0.98 for 24 requests), against 0.67 for the same variant on the first board in the run above, which shows the variance between boards and runs,} and the text-only path reaches 0.96, the variant without cancellation 0.88, waiting 0.67 (0.47 to 0.82), and a fixed deadline 0.62. \tao{The arms change several components at once, so they show the combination: cancellation, the text path, and the budget help when the link alternates or fails, and the next section shows the cost.}

\subsection{Overhead Analysis}
\noindent\textbf{Quality of the fallback models.} A smaller device model answers less accurately, and a smaller ASR model adds errors on top (Fig.~\ref{fig:e6}): the word error rate rises from 2.14\% to 2.54\% on LibriSpeech and from 9.91\% to 13.64\% on our questions with the 0.6B ASR. On 100 spoken GSM8K questions, as in VoiceBench~\cite{chen2026voicebench}, the server model with the 1.7B ASR reaches 90\%, the AGX fallback 77\%, and the smallest model with the smallest ASR 48\%, against 95\%, 82\%, and 58\% on text. \rev{Weighting each on-time answer in the simulation by the accuracy of its model (0.90 for the server, 0.77 for the AGX fallback), \tool{} scores 0.80 against 0.72 for waiting and 0.77 for the percentile hedge. }\tao{A hedged agent may return a fallback reply without notice, so the answering tier matters, and all tiers answer the same questions since benchmarks can be contaminated~\cite{chen2025benchmarking}.}

\noindent\textbf{On-device energy consumption of hedging.} The right chart of Fig.~\ref{fig:e7} shows the extra energy per request that the AGX draws over its idle power of 7.2\,W~\cite{shirvani2024duojoule}. On a good link it costs 7\,J per request, and outages raise it to 14 to 29\,J. Under periodic outages the fallback starts for 19 of the 24 requests and costs 104\,J per request, and 148\,J without cancellation, so cancellation saves 30\%.

\subsection{Practical Case Study}

\noindent\textbf{Scale-up agent with multiple high-performance GPUs.} In practice, the server side of \tool{} scales up to serve many audio agents on several GPUs~\cite{li2025lemix}, and the 7.6k-token tool prompt of an agent is repeated before every short spoken query, so a prefix cache reuses it only on the GPU that served the agent before and each GPU must maintain its own cached prefix (Fig.~\ref{fig:tool}, left). \tool{} adopts per-agent affinity, the cache-aware routing that serving systems already offer~\cite{zheng2024sglang}, and we measure its effect on the LLM stage. We served 24 such agents with Qwen3-4B on six RTX 6000 Ada GPUs \tao{of a different machine than the A4500 server of the other experiments}, one vLLM process per GPU, with 24 closed-loop clients and 96 output tokens. One GPU serves 1.3 requests per second, and the cache raises this to 3.2 at a hit rate of 57\%, as 24 prefixes exceed its KV memory. Six GPUs with round-robin routing serve 10.5 requests per second with the cache at a hit rate of 64\%, since every GPU sees every agent, while affinity keeps the hit rate at 99.9\% and serves 21.3 requests per second with a 95th percentile of 2.1\,s (Fig.~\ref{fig:tool}, right). In an open-loop run of 10 minutes at 2 to 7 requests per second, the mean time to first token is 0.84, 0.33, and 0.08\,s without the cache, with round-robin, and with affinity.

\noindent\textbf{Field study on the Singapore MRT.} The field study evaluates the policy under a real open-network trace: an iPhone 16 Pro on one 5G SIM collects the trace and the server outcomes from Tuas Link to Bugis, and \tool{} is replayed offline with measured AGX local latencies. The phone sent a spoken question every 30\,s and pinged the server every second, and the response covers the server ASR and LLM, the network, and the iOS speech rendering. Fig.~\ref{fig:field} shows on the left the RTT along the route, where the darkest color is above 1200\,ms, and on the right the first 36 minutes with 73 requests, averaged over 10\,s windows, where black ticks mark failed pings and orange bands mark the three outages with unanswered requests. The RTT has a median of 497\,ms and a 95th percentile of 1230\,ms, and 647 of 3500 pings failed in 75 outages up to 36.5\,s long. The server-only agent answers 70 of the 73 requests with a median of 6.3\,s, and \tool{} answers all 73 with the same median and 0.99 within 30\,s against 0.96, by starting the fallback for the three requests that begin while pings fail, at a 95th percentile of 19.9\,s against 12.6\,s over answered requests. \tao{The replay combines the server outcomes of the phone with the AGX local latencies, so it omits the radio and contention of a real AGX terminal, and the difference rests on three requests.}

%% file: latex/conclusion.tex
\section{Conclusion}

Audio agents on embedded devices need the server for speed and the device for availability. \tool{} runs whole stages on the server and hedges it with a complete local fallback. Its adaptive deadline matches the server on a healthy link, and on the AGX it answered every request in the evaluated outage workload, and devices that cannot host a fallback use the server path. \tao{The evaluation uses four questions and 24 requests per cell, the form rule misfires on the Nano under periodic outages, and the latency estimate goes stale when every request starts a fallback. Future work refreshes it with probe requests, conditions the form on the link state, and adds spoken questions.}